\documentclass[runningheads]{llncs}

\usepackage{graphicx}
\usepackage{latexsym}

\usepackage{mathmacros}
\usepackage{algorithm}
\usepackage{algpseudocode}

\usepackage{comment}

\usepackage{multirow}
\usepackage{subcaption}

\def\Simplex#1{\Sigma^{\left< #1 \right>}}

\begin{document}
\title{Predicate Classification Using Optimal Transport Loss in Scene Graph Generation}
\titlerunning{Predicate Classification Using Optimal Transport Loss in SGG}
%
\author{Sorachi Kurita\inst{1} \and
Oyama Satoshi\inst{2} \and
Itsuki Noda\inst{1}}

\authorrunning{S.Kurita et al.}
%
\institute{Hokkaido University \and
Nagoya City University\\
}

\maketitle              

\begin{abstract}
In scene graph generation (SGG), learning with cross-entropy loss yields biased predictions owing to the severe imbalance in the distribution of the relationship labels in the dataset. Thus, this study proposes a method to generate scene graphs using optimal transport as a measure for comparing two probability distributions. We apply learning with the optimal transport loss, which reflects the similarity between the labels in terms of transportation cost, for predicate classification in SGG. In the proposed approach, the transportation cost of the optimal transport is defined using the similarity of words obtained from the pre-trained model. The experimental evaluation of the effectiveness demonstrates that the proposed method outperforms existing methods in terms of mean Recall@50 and 100. Furthermore, it improves the recall of the relationship labels scarcely available in the dataset.
\keywords{Scene Graph Generation \and Optimal Transport \and Unbiased Training. }
\end{abstract}

\section{Introduction}
In scene graph generation(SGG), performance deteriorates due to the imbalance in the distribution of relationship labels within the dataset.\par

\begin{figure}[t]
\centerline{\includegraphics[width=\textwidth]{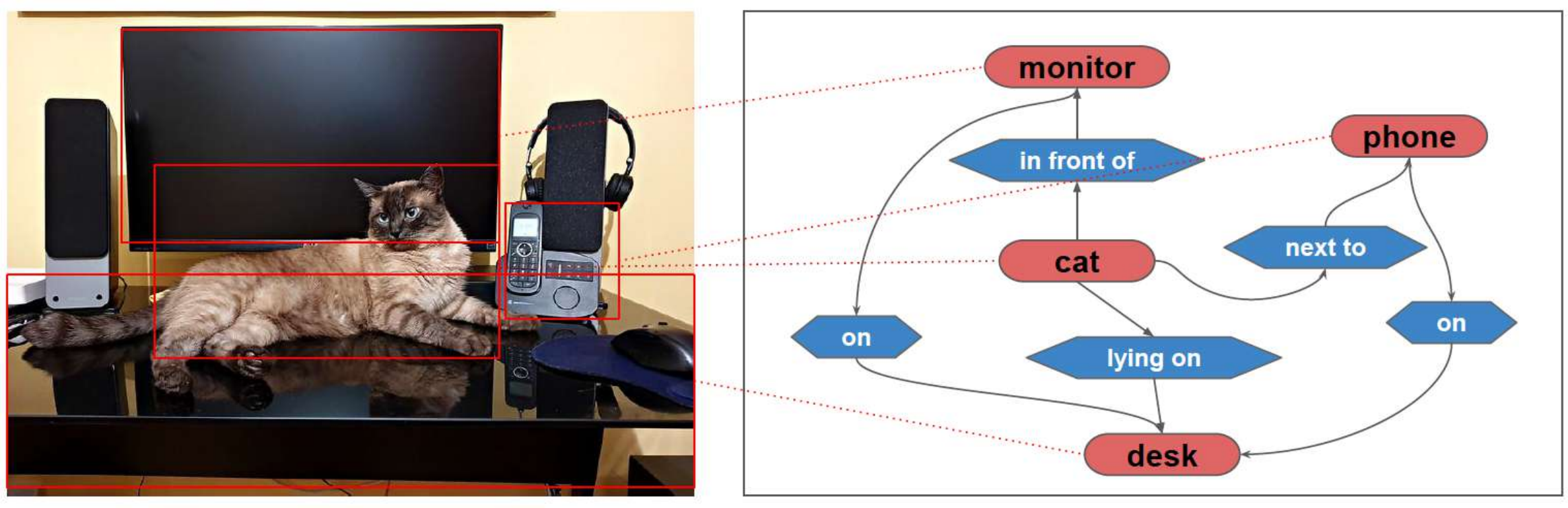}}
\caption{Scene Graph} \label{SceneGraph_Example}
\end{figure}

Recently, scene graphs have gained attention as a means of representing relationships between objects in an image, serving as an intermediary representation between language and images. A scene graph is a directed graph that succinctly represents the objects in an image and their relationships, where objects are denoted as nodes and relationships as edges, as illustrated in Fig. \ref{SceneGraph_Example}.\par

Representing the context in an image through a graph facilitates advanced tasks, including vision question answering (VQA)~\cite{VAQ_1,VQA_2}, image captioning~\cite{Image_Captioning_1,Image_Captioning_2,Image_Captioning_3} and its evaluation~\cite{Image_Captioning_eval}, image generation~\cite{Scene_Synthesis_1,Scene_Synthesis_2}, and image retrieval~\cite{Image_Retrieval_1}.

Alternatively, the gap between language and images can be bridged through the embedding of images and language into a common feature space, such as  visual-semantic embeddings (VSE)~\cite{vse,vsepp} and vision-language models~\cite{VilBERT}. The scene graphs have demonstrated usefulness in learning this approach as well~\cite{ERNIE_Vil}.

Formally, a scene graph is represented as a set of triplets $\langle subject-predicate-object \rangle$. For instance, the scene graph of Fig.~\ref{SceneGraph_Example} includes $\langle cat-lying \ on-desk \rangle$ and $\langle cat-in \ front \ of-monitor \rangle$. 

To generate a scene graph, the $\langle subject-predicate-object \rangle$ triplet is obtained by initially detecting all of the $subject$ and $object$ in the entire image, and thereafter, predicting the $predicate$ based on the information of the $subject$ and $object$. However, this stage of predicting the relationship between objects poses several challenges, listed as follows: (1) The performance of predicting relationships is worse than that of object detection, thereby causing a bottleneck in SGG. (2) Existing SGG methods frequently predict relationship labels with limited information such as ``on'' and ``in.'' (3) Inferring cases that are not included in the training data is a challenging task.

The root cause of these problems lies in the difficulty of annotating relationship labels. Generally, the terms representing a relationship between objects are not necessarily limited to a single expression. For instance,  ``sitting on'' and ``on'' are commonly used to describe a similar spatial relationship, although the former conveys a more detailed relationship than the latter. They can coexist as relationship labels, both representing the relationship between the same objects.


Herein, we propose an approach that leverages optimal transport to mitigate the bias introduced by training data in SGG. The contributions of this study, derived from this approach, are as follows:

\begin{enumerate}
\item We enhance the recall of relationship labels such as ``walking on'' and ``riding'' which contain substantial information but are less frequently encountered in the training data.
\item Consequently, our method surpasses existing methods in terms of mean Recall@50 and 100.
\end{enumerate}

To the best of our knowledge, this is the first research attempt to incorporate optimal transport into the learning process of SGG. Furthermore, the proposed approach can be seamlessly integrated into existing models without necessitating any architectural modifications.

\subsection{Notation}
\begin{itemize}
      \item An $d$-dimensional vector with all components $x$ is denoted as $\mathbb{\textbf{x}}_{d}$.
      \item Set of vectors with non-negative components and the sum of $1$ (probability simplex) is denoted as $\Simplex{d} \coloneqq \biggr\{ \textbf{x} \in \mathbb{R}^{d}_{+} \bigr|  \textbf{x}^\top \mathbb{\textbf{1}}_{d} = 1 \biggr\} $
      \item Considering the triplet of a scene graph, the term ``relationship'' is used interchangeably with $predicate$ in this paper.
\end{itemize}

\section{Related Work}
\subsection{Scene Graph Generation}

SGG denotes the process of generating a scene graph that describes the contents of an image. Typically, this involves the following steps:

\begin{enumerate}
\item Detection of all objects and their bounding boxes in the input image.
\item Prediction of relationships between pairs of objects in the image, including the type of relationship exists.
\end{enumerate}

In Step 1, object detection and bounding box prediction can be performed using models such as Faster R-CNN~\cite{Faster-RCNN}.\par

In Step 2, the prediction of object relationships is performed according to the objects and their location information obtained from Step 1. Various algorithms~\cite{IMP,Motif,VCTree,TDE_SGG} have been proposed for this stage, which is the primary focus of research in SGG. The present focus is to predict relationships between the objects in Step 2.


\subsection{Optimal Transport}
Optimal transport is a linear programming problem that involves transporting materials from one point to another while minimizing the total transportation cost.
Formally, when considering transportation from a set of points $x = \{x_{1},...,x_{n}\}$ to another set of points $y = \{y_{1},...,y_{m}\}$, with corresponding mass vectors $\textbf{a} \in \Simplex{n}, \textbf{b} \in \Simplex{m}$, the problem can be formulated as finding a transportation matrix $\textbf{P}\in \mathbb{R}^{n \times m}_{+}$ that minimizes the transportation cost, as follows:

\begin{equation}
\label{normal_ot}
    \begin{aligned}
        & \underset{\textbf{P} \in \mathbb{R}^{n \times m}_{+}}{\text{minimize}}
        & \quad \displaystyle\sum_{i=1}^{n} \displaystyle\sum_{j=1}^{m} \textbf{C}_{ij} \textbf{P}_{ij}\\
        & \text{subject to}
        & \quad \begin{array}{rcl} 
            \textbf{P} \mathbb{\textbf{1}}_{m} & = & \textbf{a}, \\
            \textbf{P}^{\top} \mathbb{\textbf{1}}_{n} & = & \textbf{b}, \\
            \end{array}
       \end{aligned}
\end{equation}

Here, $\textbf{C}$ denotes the cost matrix $\textbf{C} \in \mathbb{R}^{n \times m}$. $C_{ij}$ represents the cost of transporting one unit from point $x_{i}$ to point $y_{j}$.

The first constraint in equation (\ref{normal_ot}):
$
\textbf{P} \mathbb{\textbf{1}}_{m} = \textbf{a}
\label{ot_constraint_1}
$
represents the total amount transported from point $x_{i}$, which must be equal to the mass at point $x_{i}$.

The second constraint in equation (\ref{normal_ot}):
$
\textbf{P}^{\top} \mathbb{\textbf{1}}_{n} = \textbf{b}
\label{ot_constraint_2}
$
represents the total amount transported to point $y_{j}$, which must be equal to the mass at point $y_{j}$.

The objective function:
$
\displaystyle\sum_{i=1}^{n} \displaystyle\sum_{j=1}^{m} \textbf{C}_{ij} \textbf{P}_{ij}
\label{ot_constraint_3}
$
represents the total transportation cost.\par

This kind of formulation is called Kantorovich problem, for which an optimal solution exists at all instances.

\subsection{Sinkhorn Algorithm}

\begin{algorithm}[h]
\caption{Sinkhorn Algorithm}
\label{algo:sinkhorn_algorithm}
\begin{algorithmic}[1]
\Require $probability \ distribution\ \textbf{a} \in \Simplex{n}, \textbf{b} \in \Simplex{m},   \newline cost \ matrix \ \textbf{C} \in \mathbb{R}^{n \times m}, \ regularization \ factor \ \varepsilon$
\State $\textbf{K} \gets \exp(\frac{-\textbf{C}}{\varepsilon})$
\State $\textbf{u} \gets \mathbb{\textbf{1}}_{n}$
\State $\textbf{v} \gets \mathbb{\textbf{1}}_{m}$
\While {u has not converged}
    \State $\textbf{u} \gets \frac{\textbf{a}}{( \textbf{K} \textbf{v} )}$
    \State $\textbf{v} \gets \frac{\textbf{b}}{( \textbf{K}^\top \textbf{u} )}$
\EndWhile
\State $\textbf{P} \gets Diag(\textbf{u})\textbf{K}Diag(\textbf{v})$
\State \Return $\displaystyle\sum_{i=1}^{n} \displaystyle\sum_{j=1}^{m}\textbf{C}_{ij} \textbf{P}_{ij}$
\end{algorithmic}
\end{algorithm}

Optimal transport  can be used as a tool to compare two probability distributions, similar to the use of cross-entropy in machine learning~\cite{wasserstein}. 
One of the advantages of using optimal transport is its ability to represent the relationships and similarities between probability events as transportation costs. This is in contrast to cross-entropy, where all probability events are treated uniformly. \par
By considering optimal transport between probability distributions, the minimum transportation cost can be interpreted as the distance between them. This provides a natural mechanism to express the distance that reflects the semantic relationships between probability events.\par

However, the direct use of optimal transport in machine learning poses challenges, including computational complexity and the absence of gradient information. To address these issues, we consider the problem of optimizing the objective function of optimal transport with entropy regularization for the transportation matrix $\textbf{P}$ formulated as follows: \par

\begin{equation}
 \label{sinkhorn_ot}
    \begin{aligned}
        & \underset{\textbf{P} \in \mathbb{R}^{n \times m}}{\text{minimize}}
        & \quad \displaystyle\sum_{i=1}^{n} \displaystyle\sum_{j=1}^{m} \textbf{C}_{ij} \textbf{P}_{ij} \ &-& \varepsilon H(\textbf{P})\\
        & \text{subject to}
        & \quad \begin{array}{rcl} 
            \textbf{P} \mathbb{\textbf{1}}_{m} & = & \textbf{a}, \\
            \textbf{P}^{\top} \mathbb{\textbf{1}}_{n} & = & \textbf{b}, \\
            \end{array}
       \end{aligned}
\end{equation}

Here, the entropy of the matrix $\textbf{P} \in \mathbb{R}^{n \times m}$ is
$
H(\textbf{P}) \coloneqq -\displaystyle\sum_{i=1}^{n} \displaystyle\sum_{j=1}^{m}\textbf{P}_{ij}\log\textbf{P}_{ij}
\label{thesystem}
$
and $\varepsilon$ is a regularization factor. \par



This type of optimal transport problem with entropy regularization can efficiently determine its quasi-optimal solution using the Sinkhorn algorithm \cite{Sinkhorn}, which operates iteratively, as indicated in Algorithm \ref{algo:sinkhorn_algorithm}. Notably, rapid computation can be achieved using GPUs and this algorithm is differentiable. The worst-case computation time increased proportionally to $nm$.

\section{Approach}
Herein, we applied learning with the loss of optimal transport between the predicted and correct probability distributions, where the cost matrix is defined by the similarity of words representing the relationship obtained from BERT~\cite{BERT}, to predicate classification and examine its effectiveness.\par

\begin{figure}[h]
    \centering
    \includegraphics[width=\textwidth]{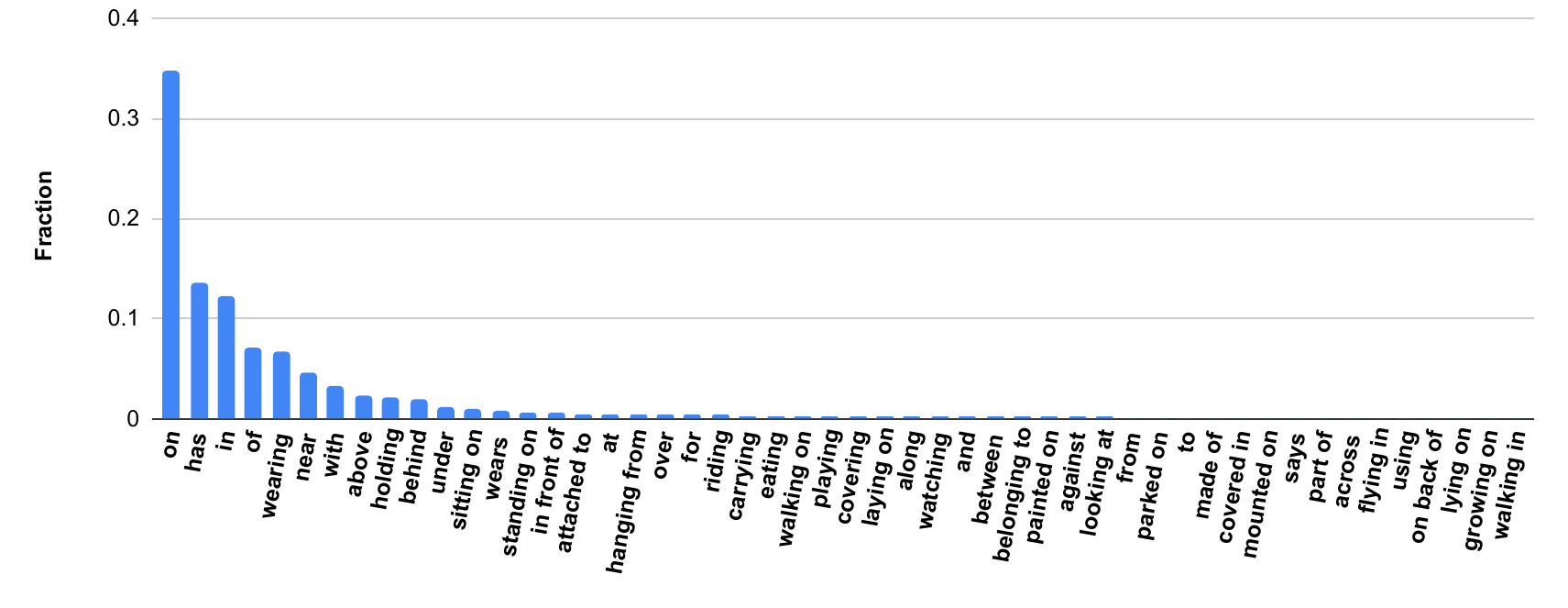}
    \caption{Distribution of each relationship label in Visual Genome~\cite{Visual_Genome} as a percentage of the whole dataset.}\label{VG_fraction}
\end{figure}

Visual Genome~\cite{Visual_Genome} is a widely used dataset for SGG, comprising 108,079 images with 75,729 object labels and 40,480 relationship labels. However, the relationship labels in this dataset exhibit a significant imbalance, as illustrated in Fig.~\ref{VG_fraction}. This imbalance can be attributed to the diverse vocabulary available for expressing relationships between objects, and the annotations of object labels and relationship labels being subject to the discretion of crowd workers.

\begin{figure}[h]
    \centering
    \includegraphics[width=\textwidth]{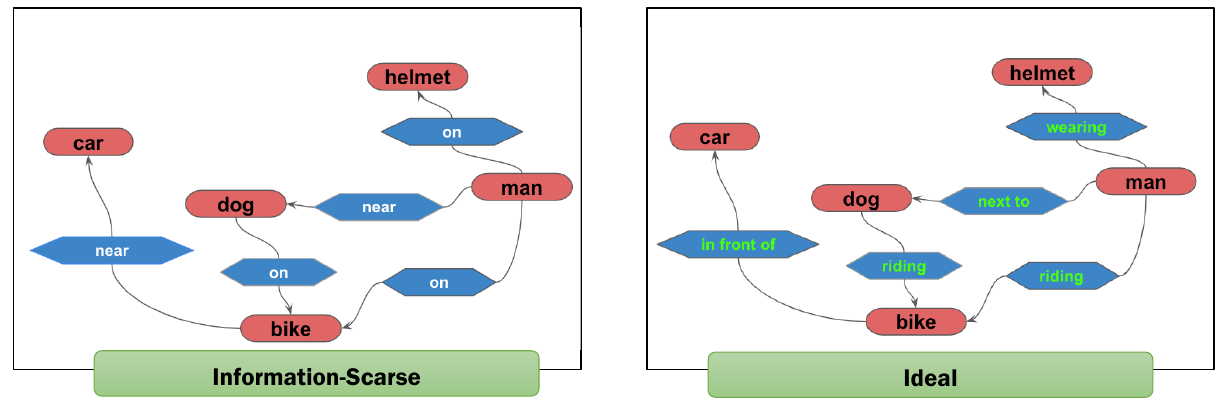}
    \caption{Information Scarce Scene Graph and Ideal Scene Graph.}
    \label{idealsg}
\end{figure}

Owing to this severe distribution imbalance, there is a higher likelihood of generating a scene graph with scarce information, as depicted on the left side of Fig.~\ref{idealsg}, rather than an ideal scene graph with rich information, as depicted on the right side.

By utilizing the optimal transport loss, our goal is to mitigate the influence of the training data distribution bias. This is achieved by allowing the misclassification of a relationship label to be shifted toward a similar relationship label based on their similarity.\par

For instance, when learning the label ``on,'' it is desirable to allow other labels with similar meanings, such as ``walking on'' or ``riding,'' to some extent. In this case, conversely, when learning ``walking on'' or ``riding,'' it is acceptable for them to be confused with ``on.'' However, since ``on'' is more frequently present in the training dataset, ``walking on'' or ``riding'' will be more tolerated than ``on'' throughout the training process.


Specifically, for $n$ relationship labels, we define the $(i,j)$ element of the cost matrix $\textbf{C}$ as follows, where $\textbf{v}_{i},\textbf{v}_{j} \quad (i,j \in \{1,2,...,n\})$ represent the feature vectors for the relationship labels corresponding to indices $i$ and $j$. These vectors are obtained from BERT~\cite{BERT}, which provides embeddings not only for typical vocabulary but also for stop words,  including ``in'' and ``of'':
$$\textbf{C}_{i,j} = 1 - \frac{\textbf{v}_{i}  \cdot  \textbf{v}_{j}}{\|\textbf{v}_{i}\|_2 \|\textbf{v}_{j}\|_2}$$
For relationship labels comprising multiple words, such as ``walking on,'' we calculate the average vector of the feature vectors of the words within that relationship label.

In SGG, if the objects share no relationship, the label ``background'' is used. The transportation cost from the ``background'' label was set to all other labels to be the maximum value among all elements in the cost matrix $\textbf{C}$, and the transportation cost from ``background'' to ``background'' was set to $0$. \par
\begin{figure}[h]
  \centering
  \begin{subfigure}[b]{0.65\textwidth}
    \includegraphics[width=\textwidth, height=0.4\textheight]{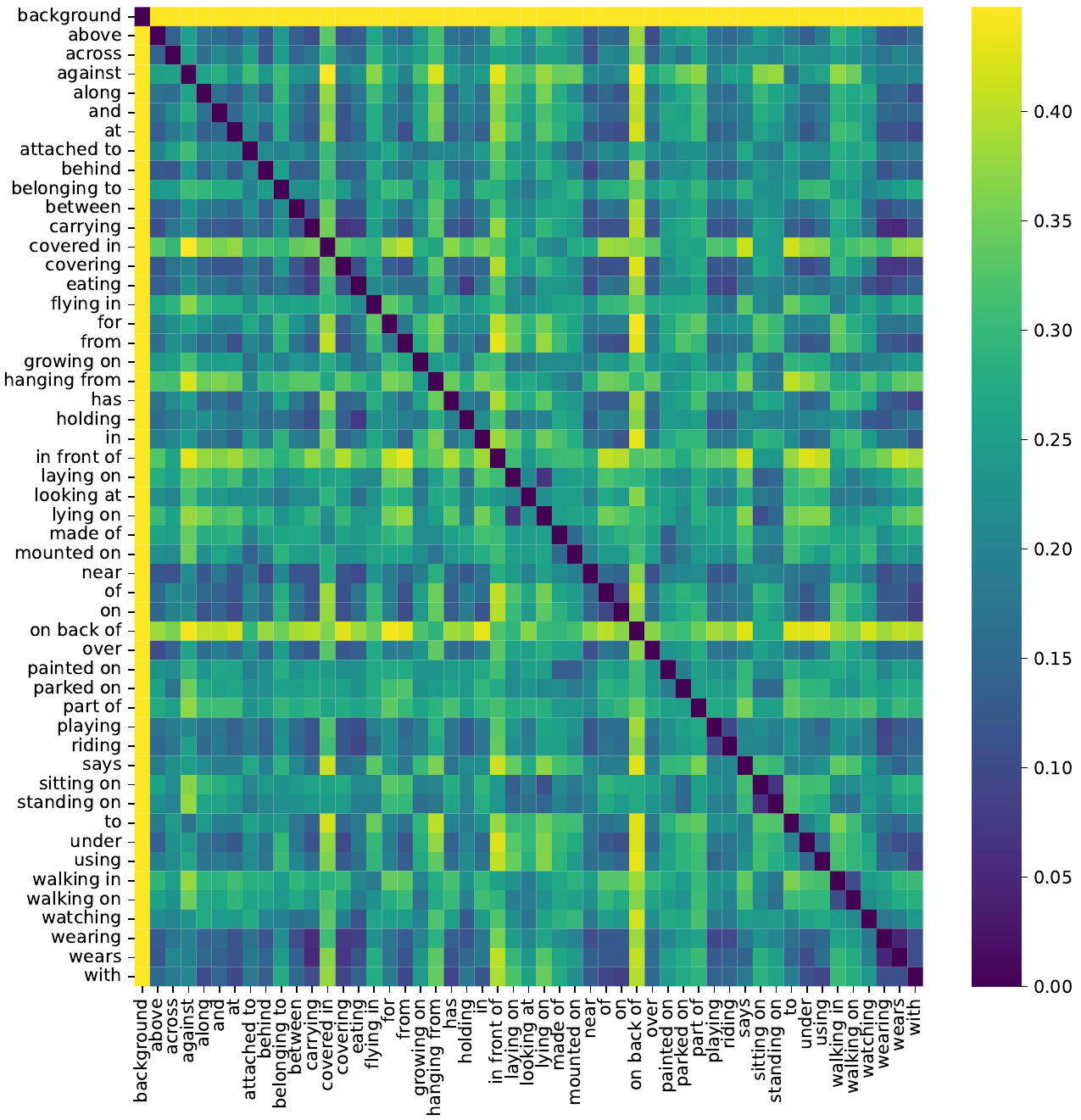}
    \caption{}
    \label{heatmap}
  \end{subfigure}%
  \begin{subfigure}[b]{0.35\textwidth}
    \includegraphics[width=\textwidth, height=0.4\textheight]{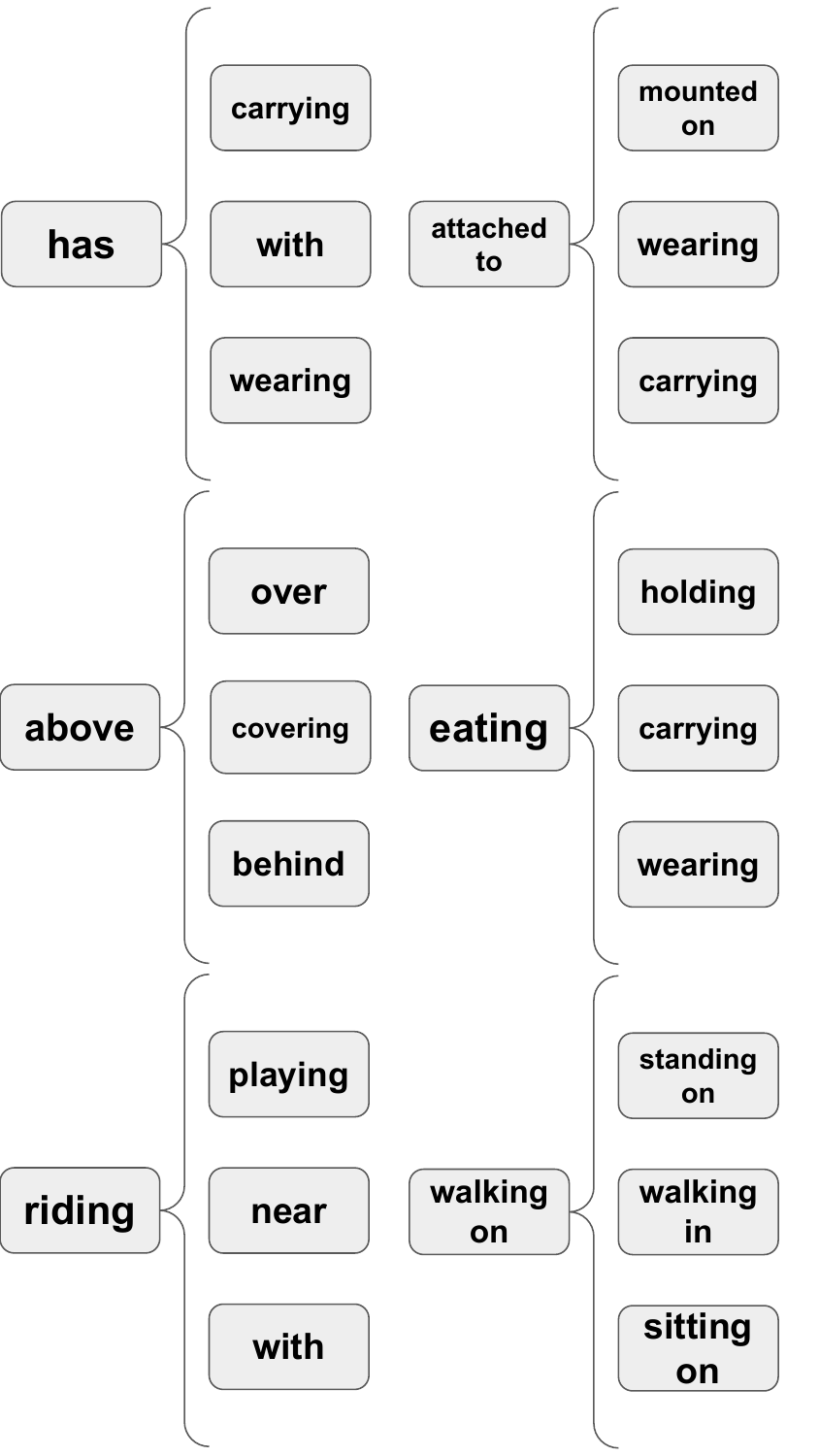}
    \caption{}
    \label{Permissible_Misclassifications}
  \end{subfigure}
  \caption{(a) \ Cost Matrix Heatmap. (b) \ Exaples of Relationship Labels and Their Corresponding Permissible Misclassifications.}
  \label{fig:two_figures}
\end{figure}

The resulting cost matrix is depicted in Fig.~\ref{heatmap}. We provide several examples of relationship labels along with the corresponding labels where misclassification is permissible based on their similarity in Fig.~\ref{Permissible_Misclassifications}.\par

Based on the obtained cost matrix $\textbf{C}$, we computed the quasi-optimal solution of the optimal transport between the predicted probability distribution $\textbf{a} \in \Simplex{n}$ and the correct probability distribution $\textbf{b} \in \Simplex{n}$ using the Sinkhorn algorithm~\cite{Sinkhorn}, as indicated in Algorithm 1. The resulting value was employed as the loss value. To ensure stability in the calculations, we adjusted the Sinkhorn algorithm in the logarithmic domain.

As discussed earlier, the relationship labels in Visual Genome~\cite{Visual_Genome}, used for learning SGG, exhibited severe imbalanced data. When learning from such imbalanced data, machine learning models with cross-entropy loss tended to favor the more frequent labels in the training data to minimize the loss within the training set. To address this issue, three general engineering methods have been proposed for existing models:

\begin{enumerate}
\item Data augmentation for classes with fewer labels (oversampling)
\item Sampling the number of data in a class with a high number of labels to match the number of data in a class with a low number of labels (undersampling)
\item Redesigning the loss function.
\end{enumerate}

Approach 1: Oversampling is the most effective and least disadvantageous method, extensively studied. However, in predicting relationships for SGG, simple data expansion techniques like affine transforms were not applicable due to the importance of positional information in the image. 


Furthermore, Approach 2: undersampling is not a practical solution because of its significant drawbacks associated with discarding data, especially considering the severe imbalance in the distribution of the relationships labels.

Therefore, for the specific case of imbalanced data, such as predicting relationships in SGG, a natural approach was to redesign the loss function, referred to as Approach 3.

In the first place, this motivation stemmed from the fact that the widely used cross-entropy loss function is highly influenced by the training data bias.

The cross-entropy used in $N$-category classification was formulated as follows, where $\textbf{p}\in \Simplex{n}$ represented the correct probability distribution and $\textbf{q}\in \Simplex{n}$ represented the predicted probability distribution$$Cross Entoropy Loss(\textbf{p},\textbf{q}) = - \displaystyle\sum_{i=1}^{n} \textbf{p}_{i}  \log \textbf{q}_{i}$$

When the correct probability distribution was provided as a one-hot label, and the index of the correct label was denoted as $t$, the loss was computed using the following equation:

$$Cross Entoropy Loss(\textbf{p},\textbf{q}) = -\textbf{p}_{t}  \log \textbf{q}_{t}$$

As observed from the above equation, the cross-entropy loss considered only the probability value of the correct label in the predicted probability distribution. Therefore, when the gradient information of this computed loss is backpropagated to the model, it becomes susceptible to biases stemming from the training data distributions.

As depicted in Fig. \ref{VG_fraction}, the training data was biased toward relationship labels with limited information, such as ``on,'' ``in,'' and ``of.'' Consequently, the learning outcomes, influenced by this bias, not only predicted more common relationship labels but struggled to predict relationship labels with richer information, such as ``walking on'' and ``riding.''

In contrast, by utilizing optimal transport as the loss function, we could incorporate the transportation of probability values from all labels to the correct label, thereby considering both the correct and non-correct labels. This approach enabled us to mitigate the impact of biased training data distribution by allowing a certain level of tolerance for relationship labels like ``walking on'' and ``riding'' when learning the label ``on.''

\section{Experiment}
\subsection{Experimental Settings}
\subsubsection{Dataset}
In this study, we adopted a modified version of Visual Genome~\cite{Visual_Genome} as the dataset, following the methodologies of previous works in this field~\cite{IMP,Motif,VCTree,TDE_SGG}. This dataset consisted of approximately 108,000 images and included the 150 object categories and 50 relationship labels with the highest data volume among those present in Visual Genome~\cite{Visual_Genome}. Similar to~\cite{Motif,TDE_SGG}, we allocated 70\% of the dataset for training, 30\% for testing, and additionally utilized a validation dataset comprising 5,000 images sampled from the training set.

\subsubsection{Scene Graph Generator}
We compared the cross-entropy loss with the optimal transport loss, as described in Section 3, by replacing the loss function in existing SGG methods, Motif~\cite{Motif} and VCTree~\cite{VCTree}. Furthermore, we evaluated the performance of IMP~\cite{IMP}, the first SGG model, as a comparison model, following ~\cite{TDE_SGG}. Although IMP was not trained using the optimal transport loss, it was considered for performance comparison with the proposed approach.\par

All of these models employed the same hyperparameters. Additionally, the value of $\varepsilon$ within the Sinkhorn algorithm was fixed at $1$.

\subsubsection{Evaluation Method}
To assess the effectiveness of the optimal transport loss, we utilized the mean Recall in Predicate Classification of Relationship Retrieval. \par
Relationship Retrieval (\textbf{RR}) consisted of the following three tasks: (1) Predicate Classification \textbf{(PredCls)}: Given the ground truth of object bounding boxes and labels, the task involved predicting the relationships between the objects. (2) Scene Graph Classification \textbf{(SGCls)}: Given the ground truth of the object bounding boxes, the goal was to predict the object labels and relationship labels. (3) Scene Graph Detection (\textbf{SGDet}): This task required predicting the object bounding boxes, object labels, and relationship labels from an image without any given ground truth.To ensure that the evaluation of the optimal transport loss was not influenced by the performance of the object detection stage in SGG, we focused on Predicate Classification (PredCls).



Lately, \textbf{mean Recall@K (mR@K)} has often been used as a metric of RR.
In the case of severely imbalanced data for relationship labels, if we can accurately predict only the labels with a large distribution, such as ``on'' or ``in,'' the \textbf{Recall@K(R@K)} is calculated to be high. This means that R@K cannot evaluate the performance of relationship labels with rich information, such as ``walking on'' and ``riding.'' 
Therefore,  in ~\cite{VCTree},  mR@K was proposed for evaluation. The mR@K was obtained by calculating the R@K for each relationship label and taking the average across all labels. We assessed the performance using mR@20, mR@50, and mR@100.

\subsection{Experimental Results}
Table~\ref{table:result} presents the performance obtained from the experiment. Each column corresponds to the SGG model mentioned in Section 4.1, the type of loss function used, and the mR@20, mR@50, and mR@100 values. In terms of the loss functions, \textbf{CE} represents cross-entropy loss, and \textbf{OT} represents optimal transport loss. The notations \textbf{OT(SUM)} and \textbf{OT(MEAN)} indicate that the loss for each batch was calculated as the sum of losses within the batch and as the average of losses within the batch, respectively.

\begin{table}[h]
  \caption{Performance Comparison}
  \label{table:result}
  \centering
  \begin{tabular}{ccccc}
    \hline
    model & Method  &  mR@20 & mR@50 & mR@100  \\
    \hline \hline
    IMP  &  CE & 9.47 & 11.82 & 12.69 \\
    \hline
    \multirow{3}{*}{Motif}  & CE & 12.53 & 15.99 & 17.30 \\
            & OT (MEAN) & 11.33 & 16.10 & 19.14 \\
            & OT (SUM)  & 12.28 & \textbf{17.55} & \textbf{20.95} \\
    \hline
    \multirow{3}{*}{VCTree}  & CE & \textbf{13.75} & 17.32 & 18.58 \\
            & OT (MEAN) & 11.42 & 16.15 & 19.24 \\
            & OT (SUM)  & 10.39 & 16.13 & 19.95 \\
    
    \hline
  \end{tabular}
\end{table}

Overall, OT(SUM) outperformed OT(MEAN) in terms of performance.
In case of Motif, the performance slightly deteriorated in mR@20, whereas it significantly improved in mR@50 and mR@100.
Similarly, in case of VCTree, the performance slightly deteriorated in mR@20 and mR@50, while it significantly improved in mR@100.
For mR@20, the conventional VCTree demonstrated the best performance, while for mR@50 and mR@100, Motif with OT(SUM) displayed the best performance.

\begin{figure}[ht]
	\centering
	\includegraphics[width=\textwidth]{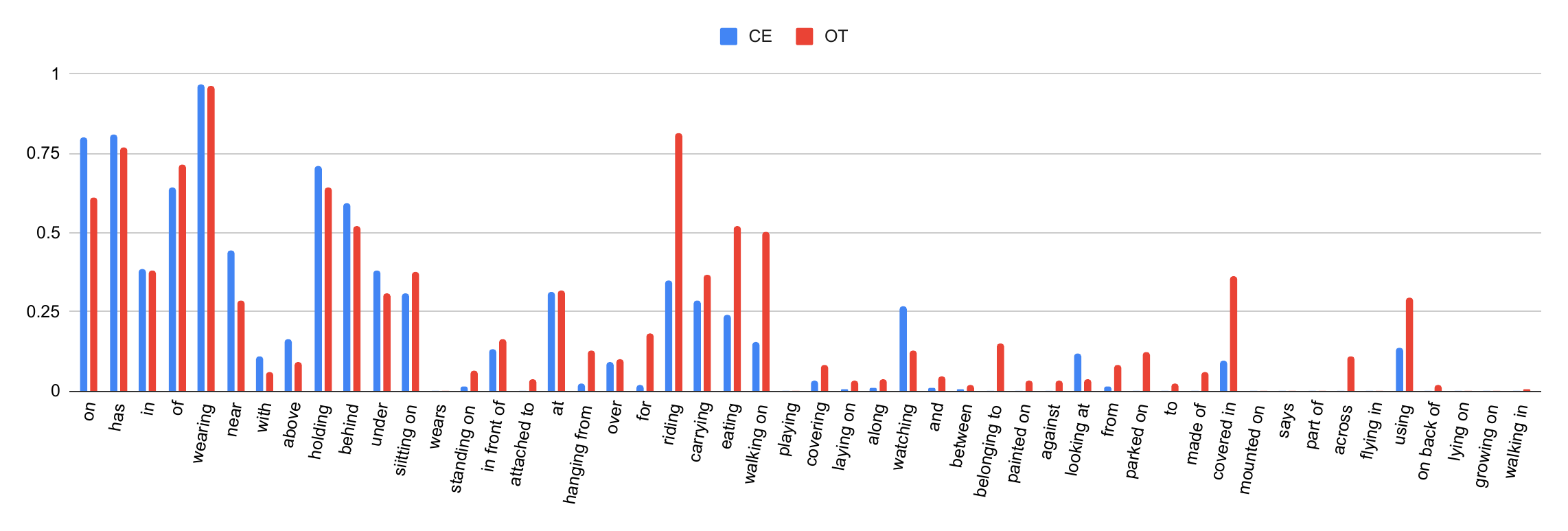}
	\caption[Recall@100 of each relationship label]{Recall@100 of each relationship label}
	\label{fig:recall.pdf}
\end{figure}

To confirm the enhancement in prediction performance of less frequent labels such as ``riding'' or ``walking on'' in the training data, the R@100 for each relationship label in Motif's CE and OT(SUM) is presented in Fig. \ref{fig:recall.pdf}. In particular, CE was trained with cross-entropy loss and OT was trained with OT(SUM). From left to right, the horizontal axis lists the relationship labels that were more frequently observed in the training data. The vertical axis represents R@100 for each relationship label.

The relationship labels present on the left-hand side, which were more frequent in the training data, exhibited a slightly worse performance. In contrast, the performance on the right-hand side of the figure improved as well, especially for ``riding,'' ``eating,'' and ``walking on'' labels. Moreover, the proposed method can predict labels such as ``belonging to,'' ``parked on,'' and ``across'' that could not be predicted earlier using the conventional method.

\subsection{Discussion}
The optimal transport loss yielded a loss design considering non-correct labels based on their similarity, unlike learning with one-hot correct labels and cross-entropy loss. Thus, regarding Table~\ref{table:result}, the predicted probability values of the correct labels were lower, which potentially degraded the performance of mR@20 in Motif and mR@50/100 in VCTree.


Regarding Fig. \ref{fig:recall.pdf}, it is believed that the use of optimal transport loss can mitigate the impact of bias in relationship labels, resulting in diversified predicted relationship labels. 


\section{Conclusion}
In the SGG, a significant imbalance exists in the relation labels of the training data. This results in a problem where learning using conventional cross-entropy loss tend to produce predictions biased toward this distribution. To address this issue, we applied the optimal transport loss, where the transport cost was defined based on the similarity of relationship labels. This enabled learning that tolerated mistakes among semantically similar labels and yielded the following results. \par

\begin{enumerate}
\item The proposed method enhanced the recall of relationship labels such as ``walking on'' and ``riding,'' which contain a large amount of information but appear less frequently in training data.
\item The proposed method outperformed existing methods in terms of mean Recall@50 and 100.
\end{enumerate}
\par

Future research should focus on refining the optimal transport loss, such as extracting better feature representations of relationship labels.\par
We anticipate that these results will improve the performance of various advanced tasks as mentioned in Section 1.

%
%
%
\bibliographystyle{splncs04}
\bibliography{main}

\end{document}